\documentclass[letterpaper, 10 pt, conference]{ieeeconf}  

\IEEEoverridecommandlockouts                              

 \usepackage{amsfonts}

\usepackage{cite}
\usepackage{graphicx}    
\usepackage{booktabs}    
\usepackage{array}        
\usepackage{multirow}     
\usepackage{amsmath}     
\usepackage[section]{placeins}      
\usepackage{hyperref}
\usepackage{cleveref} 

\title{\LARGE \bf
SemSight: Probabilistic Bird's-Eye-View Prediction of Multi-Level Scene Semantics for Navigation
}

\author{Jiaxuan He$^{1}$, Jiamei Ren$^{2}$, Chongshang Yan$^{3}$, Wenjie Song$^{*}$
\thanks{*This work was partly supported by Program for National Natural Science Foundation  of  China  (Grant  No.  62373052),  Beijing  Natural  Science Foundation (Grant No. 4252051), and in part by the National Key Laboratory of Science and Technology on Space Born Intelligent Information Processing TJ-01-22-09.}
\thanks{The authors are with the School of Automation, Beijing Institute of Technology, Beijing 100081, China, (Corresponding author: Wenjie Song, email: songwj@bit.edu.cn).}
\thanks{Dataset available at: \url{https://anonymous.4open.science/r/rplan_npz-F1C5}}
}
\begin{document}

\maketitle
\thispagestyle{empty}
\pagestyle{empty}


  

\begin{abstract}

In target-driven navigation and autonomous exploration, reasonable prediction of unknown regions is crucial for efficient navigation and environment understanding. Existing methods mostly focus on single objects or geometric occupancy maps, lacking the ability to model room-level semantic structures. We propose SemSight, a probabilistic bird’s-eye-view prediction model for multi-level scene semantics. The model jointly infers structural layouts, global scene context, and target area distributions, completing semantic maps of unexplored areas while estimating probability maps for target categories. To train SemSight, we simulate frontier-driven exploration on 2,000 indoor layout graphs, constructing a diverse dataset of 40,000 sequential egocentric observations paired with complete semantic maps. We adopt an encoder–decoder network as the core architecture and introduce a mask-constrained supervision strategy. This strategy applies a binary mask of unexplored areas so that supervision focuses only on unknown regions, forcing the model to infer semantic structures from the observed context. Experimental results show that SemSight improves prediction performance for key functional categories in unexplored regions and outperforms non-mask-supervised approaches on metrics such as Structural Consistency (SC) and Region Recognition Accuracy (PA). It also enhances navigation efficiency in closed-loop simulations, reducing the number of search steps when guiding robots toward target areas.

\end{abstract}

\section{Introduction}

Autonomous robot navigation requires not only local perception but also the ability to predict unobserved areas. However, occlusions and limited sensing range often force robots to rely on egocentric partial observations. As a result, they lack a holistic understanding of the environment’s structural and semantic layout. Critical semantic structures in unobserved regions (e.g., corridors or doorways) have a notable impact on the efficiency of navigation. Therefore, predicting unknown areas from partial observations is a key challenge. The goal is to infer potential traversable structures and semantic distributions that support efficient exploration and navigation.

\begin{figure}[htbp]
  \centering
  \includegraphics[height=\columnwidth, angle=-90, keepaspectratio]{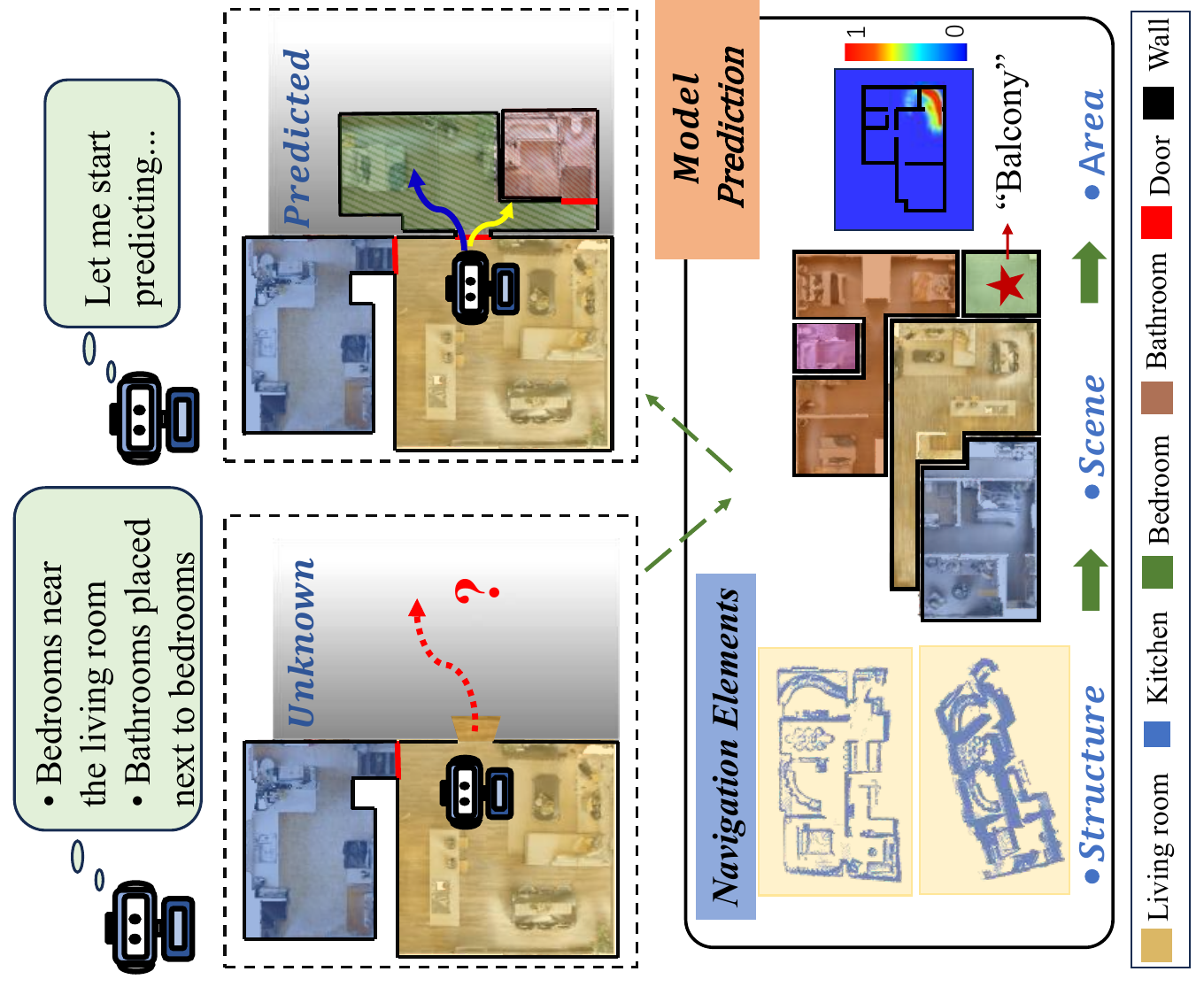} 
  \caption{With partial observations, robots cannot form a complete understanding of the environment. SemSight proposes a method for predicting the structural-semantic layout of unobserved areas in unknown environments using available information. This allows robots to “imagine the unseen world” and formulate navigation plans more efficiently.}
  \label{fig:figure1}
  \vspace{-0.3cm}
\end{figure}

Recent studies have explored scene reasoning and map prediction, employing learning-based approaches to infer unknown regions from partial observations. Most existing work focuses on geometric occupancy completion\cite{shrestha2019learned,ramakrishnan2020occupancy,wang2021learning,pronobis2017learning,pronobis2017deep} and fails to address higher-level semantic reasoning. For example, such methods cannot answer questions like “where is the bedroom or kitchen likely to be?”. Other approaches target specific semantic categories\cite{liang2021sscnav,chaplot2020object,georgakis2021learning,chang2020semantic,ji2024diffusion} but cannot capture semantic regularities across large spatial extents. As a result, their predictions are often fragmented and localized, without providing room-level semantic priors. High-level semantic structures play an important role in autonomous navigation. For instance, in target-driven tasks, a robot that anticipates “an unknown region is likely a bedroom” or “a corridor connects to other rooms” can plan paths more efficiently and accomplish navigation more effectively. Such semantic prediction capability provides a valuable prior for task-level reasoning. Compared with robots that rely solely on immediate perception, robots capable of scene-structural reasoning and semantic prediction gain the ability to “imagine the unseen world.”

Motivated by this, we propose \textbf{SemSight}, a probabilistic \textbf{sem}antic fore\textbf{sight} model. It is designed to predict the global semantic structure of an environment from partial observations, rather than being limited to local occupancy estimation. SemSight takes sparse exploration observations and target categories as input, predicts probabilities associated with target-related structures, and infers room-level semantic layouts and structural relations. By formulating semantic map prediction as a probabilistic inference problem, we directly support structured reasoning for navigation decisions, going beyond pure geometric reconstruction.

To train the SemSight model, we construct a semantic prediction dataset under the BEV representation. Unlike previous methods relying on static annotations, we simulate boundary-driven exploration processes on a large collection of floorplans, This process dynamically collects egocentric observation sequences paired with complete semantic maps as training samples. This dataset covers diverse indoor layouts with rich room types and structural information, effectively supporting the learning of local-to-global semantic reasoning. In summary, the main contributions are as follows:

\begin{itemize}
\item \textbf{SemSight model} – a probabilistic prediction model that integrates \textit{structural layouts, global scene context}, and \textit{target area distributions} for navigation. It maps partial observations and structural target conditions to global semantic maps and target probability maps, supporting target-driven navigation and exploration.
\item \textbf{Diverse boundary-driven exploration dataset} – establishing 40,000  pairings of continuous partial observations and complete semantic maps, covering 2,000 varied indoor layouts. This provides diverse training data for semantic completion and prediction tasks.
\item \textbf{Semantic foresight with navigation impact} – experiments demonstrate that our mask-constrained supervision enables SemSight to infer room-level layouts and probabilistic target area distributions in unexplored regions. Furthermore, navigation simulations show that incorporating SemSight predictions improves exploration efficiency and target finding success, confirming its practical value for autonomous navigation.
\end{itemize}

\section{Related Works}

\subsection{Map Prediction and Scene Reasoning}

Map inpainting and completion aim to predict the missing parts of a map, thereby inferring information beyond the robot’s perceptual range. Traditional approaches primarily rely on geometric modeling\cite{shrestha2019learned,ramakrishnan2020occupancy,elhafsi2020map,mann2022predicting,schreiber2019long}. For example, variational autoencoders (VAEs) have been used to predict unknown areas in occupancy grid maps\cite{shrestha2019learned}, enhancing the efficiency of robotic exploration. The occupancy anticipation model\cite{ramakrishnan2020occupancy} directly learns from first-person RGB-D observations to infer future geometric occupancy states. Although these methods demonstrate strong performance in geometric occupancy prediction, they lack the ability to anticipate high-level semantic information.

To address this limitation, semantic scene completion\cite{shah2025foresightnav,wang2024voxel,chen2024sepaint,liang2021sscnav} advances the task further. For instance, SePaint\cite{chen2024sepaint} employs a multinomial diffusion model to inpaint missing regions in BEV semantic maps, ensuring semantic consistency. However, this method mainly relies on local semantic context for repair and produces a single deterministic completion result, without providing uncertainty estimation for the prediction.

In contrast, our SemSight model advances beyond deterministic completion by performing probabilistic semantic reasoning in the BEV space, enabling structured and uncertainty-aware predictions over large unexplored regions.

\subsection{ObjectGoal Navigation and Target Prediction}

In Object Goal Navigation (ObjectNav) tasks, a robot is required to locate a specific object category (e.g., “find a chair”)\cite{deitke2022️,wijmans2019dd,ramakrishnan2022poni,zhu2022navigating,zhang2024imagine,marza2022teaching}. Since the environment is often unknown and only partially observable, relying solely on instantaneous perceptual inputs makes efficient exploration difficult. To address this limitation, recent studies have gradually introduced prediction mechanisms\cite{Ginting2024SEEKSR,song2024p2} into navigation tasks. For example, some works use self-supervised learning to predict amodal room instances in unobserved regions\cite{narasimhan2020seeing}, thereby completing room layouts. However, these methods focus only on room instance boundaries and layouts, lacking semantic prediction of room categories. SSCNav\cite{liang2021sscnav} employs a confidence-aware semantic scene completion approach to construct global 3D semantic maps for navigation. However, its outputs are deterministic geometric completions rather than probabilistic predictions for specific targets. PEANUT\cite{zhai2023peanut} uses a semantic segmentation model to predict the potential distribution of target objects in unobserved regions, which is similar to our work. However, unlike PEANUT, our SemSight model performs room-level (high-level) semantic reasoning in the global BEV space. It also provides structured probabilistic prediction, enabling more informative semantic priors.

\section{Problem Formulation}

We define the semantic prediction problem for robot navigation in unknown indoor environments as follows:

During exploration, the robot receives a sequence of egocentric observations 
$\mathcal{O} = \{o_1, o_2, ..., o_T\}$. 
At each time step $t$, the observation $o_t$ contains:  
the robot’s current position $p_t \in \mathbb{R}^2$,  
its historical trajectory $H_t = \{ p_{\tau} \mid \tau < t,\ p_{\tau} \in \mathbb{R}^2 \}$,  
the explored-region mask $E_t \in \{0,1\}^{H \times W}$ (1 denotes explored areas and 0 denotes unexplored),  
the obstacle mask $B_t \in \{0,1\}^{H \times W}$ (1 denotes occupied cells and 0 denotes free space),  
and a local multi-channel semantic map $M_t^{\text{local}} \in \mathbb{R}^{H \times W \times C}$,  
where $H \times W$ is the map resolution and $C$ is the number of semantic categories (e.g., rooms, doors, walls).  

Given a query $q$ from the room-structure layer (e.g., bedroom), 
the model learns a mapping function

\begin{equation} \label{eq:loss} 
\mathcal{F}: (o_t, q) \;\mapsto\; (\hat{M}_t^{\text{global}}, P_t^q),
\end{equation}

\noindent where $\hat{M}_t^{\text{global}} \in \mathbb{R}^{H \times W \times C}$ is the completed global semantic map, 
and $P_t^q \in \mathbb{R}^{H \times W}$ is a probability map corresponding to the query category $q$.  
Each element $P_t^q(i,j) \in [0,1]$ represents the probability that the pixel at location $(i,j)$ 
belongs to the queried semantic category $q$.

\section{Method}
\label{sec:method}

\begin{figure*}[ht]
  \centering
  \includegraphics[width=1\textwidth]{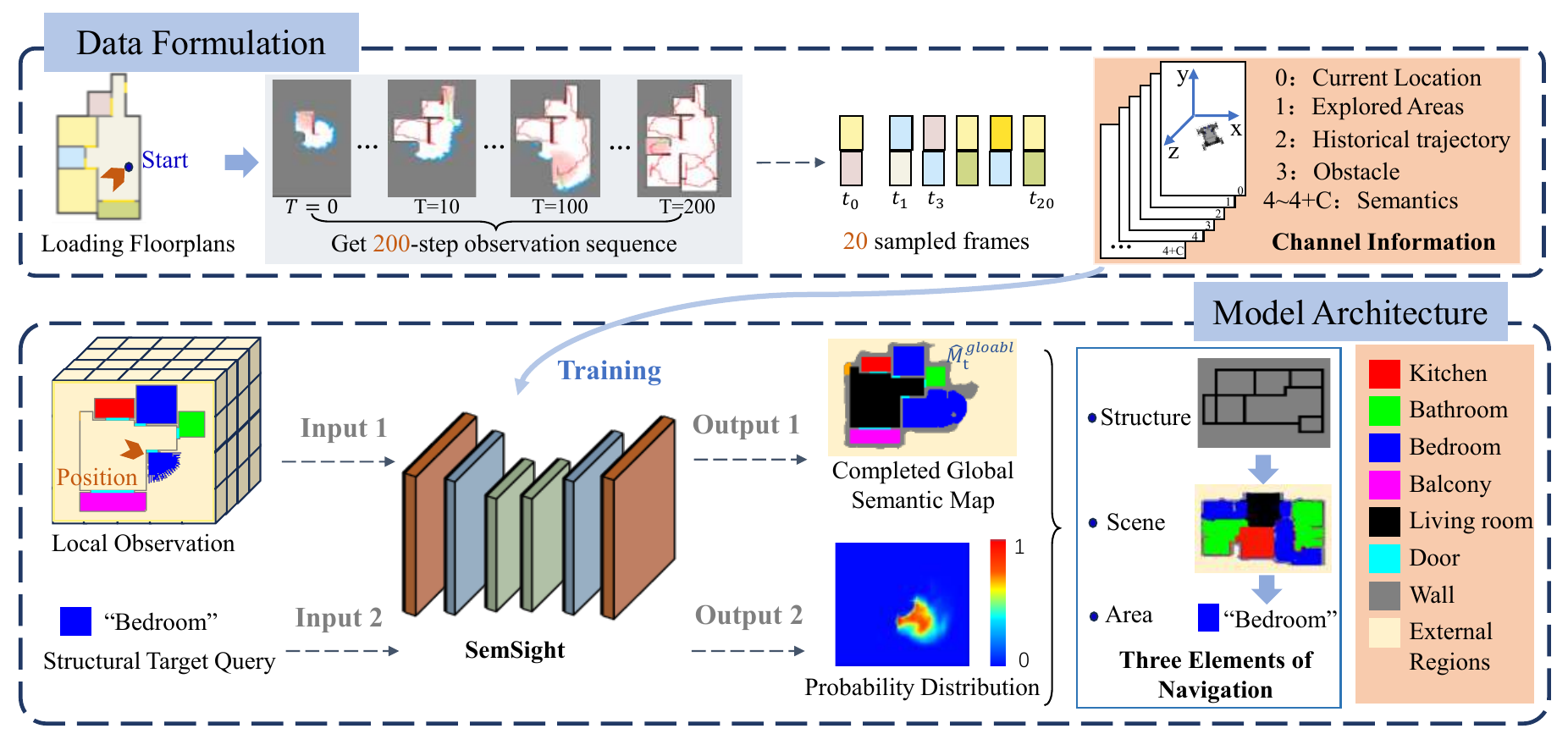} 
  \caption{\textbf{An overview of our framework for room-level semantic and target area prediction.} (a) Dataset Construction Pipeline: the simulation of frontier-driven exploration in diverse indoor layouts. The process generates paired training data of sequential observations and semantic maps. (b) SemSight Model Architecture: the encoder–decoder network used for semantic prediction. The model takes a partial observation and a target area query as input, and produces a complete semantic map together with a target probability heatmap.}
  \label{fig:figure2}
\end{figure*}

\begin{figure}[htbp]
  \centering
  \includegraphics[width=1\columnwidth]{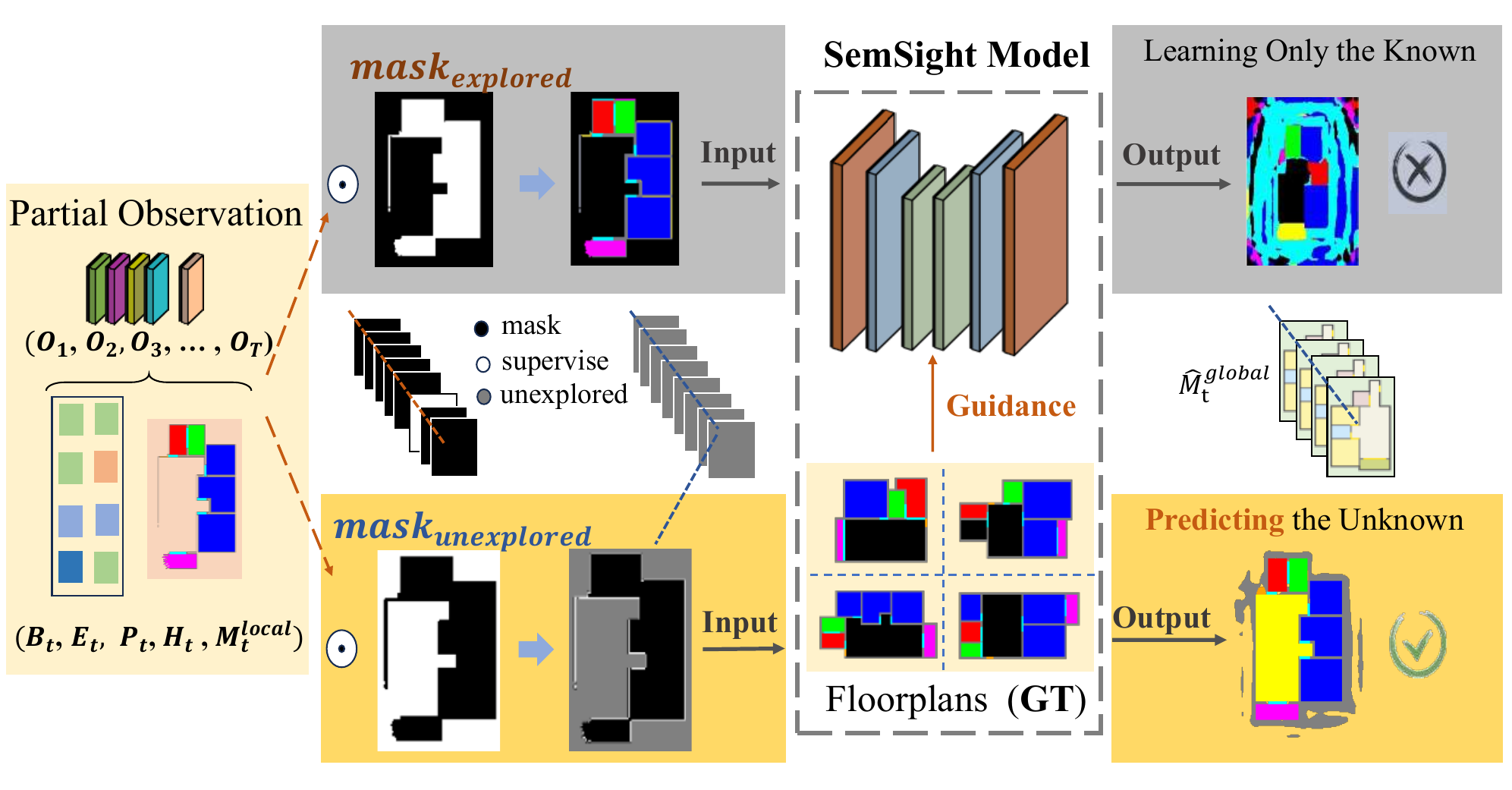} 
  \caption{\textbf{Comparison of masking strategies.}
Top: Applying supervision in explored regions leads the model to learn a trivial “copy” behavior.
Bottom: Ignoring explored regions and supervising only unexplored areas provides a clear learning signal. 
}
  \label{fig:figure3}
\end{figure}

In this section, we present our proposed probabilistic bird’s-eye-view (BEV) framework for structured scene-semantic and target area prediction (as illustrated in Fig. \ref{fig:figure2}). The core objective is to infer the probability distribution of specified target areas in unexplored regions, while also predicting the global room-level semantic layout. The framework consists of two main components: (I) Data Formulation. This module simulates robotic exploration to build a training dataset. It generates paired data of sequential egocentric observations and corresponding ground-truth global semantic maps. (II) Probabilistic Semantic Prediction with SemSight. This module applies the SemSight model to produce a complete BEV semantic map. It also outputs spatial probability maps for specified target areas, enabling structured and uncertainty-aware reasoning for navigation tasks.

\subsection{Data Formulation}

\noindent\textbf{Dataset.} We build the training samples for semantic prediction based on the RPLAN dataset\cite{wu2019data}. This dataset contains a large number of residential floor plans that are manually designed and annotated. Each floor plan is decomposed into multiple channels, including semantic labels, boundary information, room instance channels, and obstacle information. In its original definition, RPLAN provides a fine-grained categorization of room types (e.g., bedroom, kitchen, bathroom, living room). It also includes additional channels for walls, doors, and outside areas. To fit our prediction task, we reorganize the dataset into \textbf{10 semantic channels}: \textbf{7 room classes} (bedroom, living room, kitchen, bathroom, balcony, storage, doorway), \textbf{walls} (all wall segments and other impassable boundaries), \textbf{entrance doors} (unit/entry doors that connect the dwelling to the outside or public corridors, excluding interior room doors), and \textbf{the outside area}.

This reorganization preserves the essential room semantics while simplifying the label space. For example, multiple bedroom subtypes in the original dataset (e.g., master bedroom, children’s room, elderly room) are merged into a single “bedroom” category, reducing unnecessary redundancy. In addition, we explicitly model walls, entrance doors, and outside areas. This ensures that the constructed dataset maintains completeness at the structural, semantic, and scene levels.

\noindent\textbf{Exploration Dataset Construction.}To simulate the robot’s exploration process in unknown environments, we adopt a frontier-based exploration algorithm\cite{9636575} on RPLAN floorplans. At each timestep $t$, the robot receives an observation $o_t$, which includes:

\begin{itemize}

\item Positional information: the current position $p_t$ and its historical trajectory $H_t$;

\item Accessibility and occupancy information: the explored-region mask $E_t$ and the obstacle mask $B_t$;

\item Local semantic observation: a semantic map $M_t^{\text{local}}$ consisting of the 10 semantic channels described above.

\end{itemize}

During data storage, we sample the first $20$ frames from each exploration sequence. These frames are paired with the complete global semantic map $M^{\text{gt}}$, which serves as the supervision signal. In this way, each training sample is represented as ${(o_t, M^{\text{gt}})}$. This setup enables the model to learn the global semantic distribution and generate probability maps for queried categories from partially observed inputs.

\subsection{Probabilistic Semantic Prediction with SemSight}

\noindent\textbf{Network Structure.} We propose SemSight, a probabilistic semantic prediction model built upon the PSPNet\cite{zhao2017pyramid} architecture. It predicts the probability of specified structural categories and the global room-level semantic distribution from partial observations. We formulate the task as dense pixel-wise prediction and train the network to perform semantic completion. Specifically, we modify the PSPNet input to $4 + C$ channels. The first four channels encode structural information, including robot position $p_t$, exploration history $H_t$, obstacles $B_t$, and explored regions $E_t$. The remaining $C$ channels represent semantic information.

\noindent\textbf{Loss Function.} We design a multi-task weighted loss function to jointly optimize the global semantic map completion and target-class probability prediction tasks:

\begin{equation} \label{eq:loss} 
L = \lambda_{\text{global}} \cdot L_{\text{BCE}}(\hat{M}_t^{\text{global}}, M_t^{\text{gt}}) + \lambda_{\text{area}} \cdot L_{\text{BCE}}(P_t^q, Y_t^q),
\end{equation}

\vspace{6pt}
\noindent where $\hat{M}_t^{\text{global}}$ denotes the predicted global semantic map, $P_t^q$ represents the predicted spatial probability distribution of the queried target class $q$, $M_t^{\text{gt}}$ is the corresponding ground-truth semantic map, and $Y_t^q$ is the binary ground-truth mask of the target class for the unobserved region (1 if the pixel belongs to class $q$, and 0 otherwise). $\lambda_{\text{global}}$ and $\lambda_{\text{area}}$ are hyperparameters balancing the two tasks.

The weighted binary cross-entropy (BCE) loss is defined as:

\begin{equation}
\begin{aligned}[t]
L_{\text{BCE}}(x,y) = & -\frac{1}{N} \sum_{c=1}^{C} w_c \sum_{i,j} \Big[ y_{i,j,c} \log \sigma(x_{i,j,c}) \\
                     & + (1-y_{i,j,c}) \log (1-\sigma(x_{i,j,c})) \Big],
\end{aligned}
\end{equation}
 
\noindent where $\sigma(\cdot)$ is the Sigmoid function, $w_c$ is the class-specific weight assigned to structure category $c$ to handle class imbalance, and $N$ is the number of valid pixels.

\noindent\textbf{Training Strategy.}
We adopt a mask-constrained supervision strategy to guide the model in predicting semantic structures of unexplored regions from partial observations. Specifically, we construct a binary mask of explored areas, denoted as $mask_{\text{explored}}$. The supervision signal is obtained by multiplying the ground-truth semantic map $M_t^{\text{gt}}$ with the complement of this mask, ensuring that only labels in unexplored regions contribute to the loss:

\begin{equation}
\begin{aligned}[t]
\tilde{M}_t^{\text{gt}} = M_t^{\text{gt}} \odot (1 - \text{mask}_{\text{explored}}),
\end{aligned}
\end{equation}

\noindent where explored pixels are assigned zero weight in the loss function. This ensures that the model focuses on predicting the semantic distribution of unexplored regions based on explored observations.

As illustrated in Fig. \ref{fig:figure3}, different masking strategies lead to distinct supervision effects. If the unexplored regions are masked, the supervision signal is removed entirely. In contrast, masking the explored regions and retaining only the unexplored ground truth provides consistent supervision. This design encourages the model to complete missing structures and generate coherent semantic layouts.

\section{Experiments}

\subsection{Experimental Setup}

\noindent\textbf{Dataset.} Our experiments are conducted on a diverse simulated dataset constructed following the pipeline in Section \nameref{sec:method}. The dataset includes 2,000 indoor layouts, each with 200 exploration steps simulated by a frontier-based algorithm, yielding about 40,000 pairs of egocentric observations and global semantic maps. For training, we use only the first 10 frames of each sequence. These frames contain semantic maps, occupancy maps, robot poses, and trajectories. Restricting supervision to early steps ensures that the model learns to infer unexplored regions without being biased by near-complete exploration states.

\begin{figure*}[ht]
  \centering
  \includegraphics[width=1\textwidth]{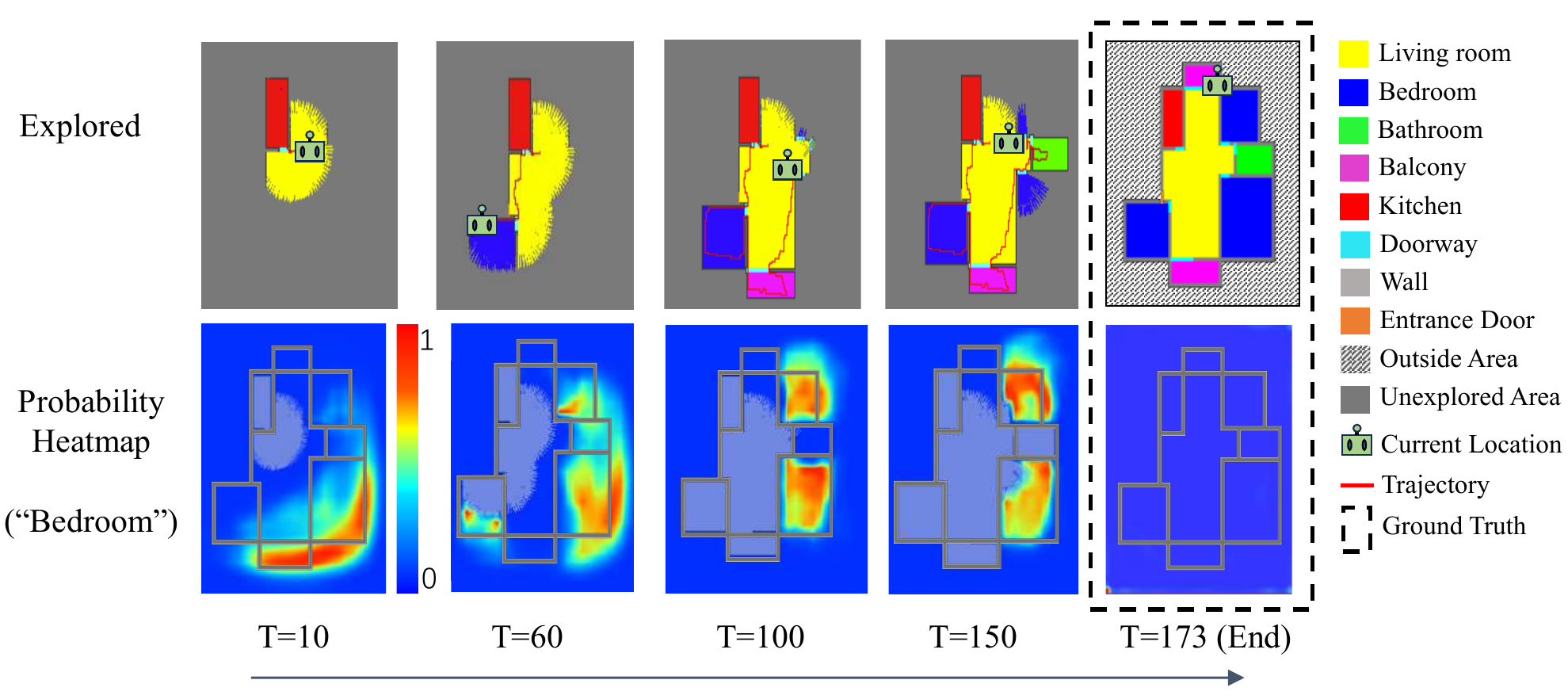} 
  \caption{\textbf{Temporal Evolution of Target Area Probability Heatmaps.} At the early stage of exploration (step 10), limited observations produce almost random predictions. As more structural and semantic cues are revealed, the model gradually refines its inference of the bedroom location. By the end of exploration (step 173), when the environment is fully observed, the model stops predicting and the heatmap probabilities diminish toward zero.}
  \label{fig:figure4}
\end{figure*}

\begin{figure}[ht]
  \centering
  \includegraphics[width=1\columnwidth]{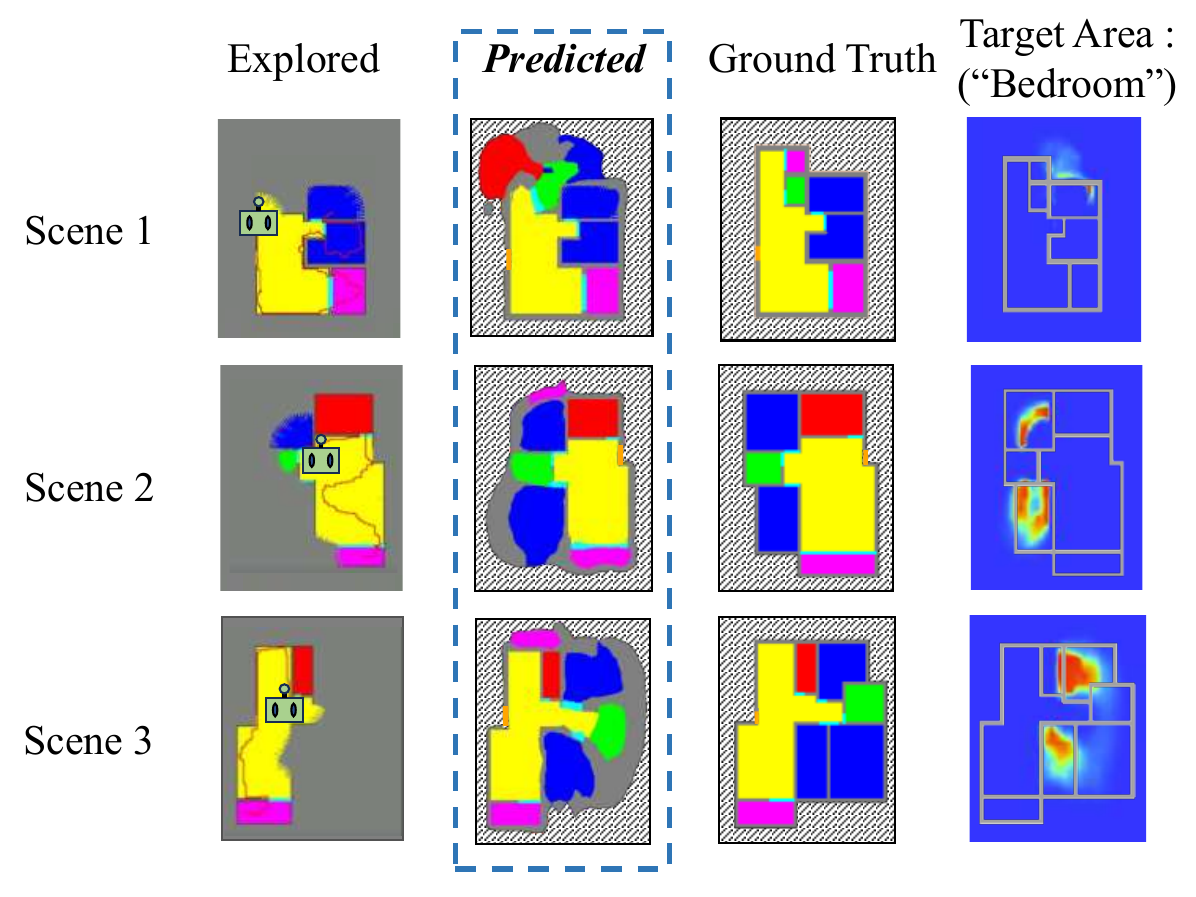} 
  \caption{\textbf{Prediction results for typical scenarios with bedroom as the target category.} Each group of results shows, from left to right: the input exploration map, the predicted semantic map, the ground-truth semantic map, and the probability heatmap for the target category (bedroom).
}
\vspace{-0.3cm}
  \label{fig:figure5}
\end{figure}

\noindent\textbf{Training Details.} The model was trained with the MMSegmentation framework\cite{contributors2020mmsegmentation}, using a ResNet50 encoder\cite{he2016deep} and a PSPNet decoder\cite{zhao2017pyramid}. The input has 14 channels, combining exploration states with structural semantic information. Training used the Adam optimizer\cite{Kingma2014AdamAM} with an initial learning rate of $5\times10^{-4}$, and a polynomial decay schedule (power = 0.9, minimum $1\times10^{-5}$) for stability.

During training, data augmentation includes random rotation (range $[-180^\circ,180^\circ]$), random horizontal flipping, random cropping ($256\times256$), and padding. The model was trained for 60,000 iterations with a batch size of 8 on a single NVIDIA RTX A6000 GPU. Validation and evaluation were performed every 2,000 iterations.

\subsection{Quantitative Results.}

\noindent\textbf{Evaluation Metrics.} Unlike standard semantic segmentation, our task focuses on predicting global layouts and target area distributions in unexplored regions. For better evaluation, we adapt conventional metrics to emphasize region-level plausibility rather than pixel-level precision. Specifically, we report \textbf{FWIoU} and \textbf{PA} within unexplored regions, while relaxing the requirement for strict boundary alignment. We also evaluate target categories using recall, precision, and F1-score to capture whether functional areas are correctly identified. In addition, we adopt a Structural Consistency \textbf{(SC)} metric to assess the plausibility of predicted room relationships and connectivity.

\noindent\textbf{Necessity of Mask-Constrained Supervision.} To assess the necessity of our supervision strategy, we compare against an observed-only PSPNet baseline, where loss is computed solely on explored pixels as an adaptation of conventional semantic segmentation. The results show that the baseline almost entirely fails to predict unexplored regions, yielding only 25.5\% FWIoU, 26.3\% PA, and 12.6\% SC. Its predictions tend to replicate visible patterns while lacking inference for unknown regions. In contrast, SemSight achieves 69.3\% FWIoU, 86.1\% PA, and 64.8\% SC under mask-constrained supervision, where supervision is explicitly applied to unknown regions. These findings confirm that specialized supervision is indispensable for semantic foresight in unexplored regions.

\vspace{-3pt}
\begin{table}[htbp]
\centering
\caption{Performance on Key Categories}
\label{tab:metrics}
\begin{tabular}{lccc}
\toprule
\textbf{Class}      & \textbf{Recall $\uparrow$} & \textbf{Precision $\uparrow$} & \textbf{F1-score $\uparrow$} \\
\midrule
Bathroom   & 70.9\%    & 68.6\%      & 69.7\%     \\
Kitchen    & 66.3\%    & 64.2\%      & 65.2\%     \\
Bedroom    & 80.2\%    & 82.9\%      & 81.5\%     \\
Living Room & 86.5\%   & 88.9\%      & 87.7\%     \\
\bottomrule
\end{tabular}
\end{table}

\noindent\textbf{Key Category Performance Analysis.} As shown in Table \ref{tab:metrics}, our model performs particularly well on large-area categories such as Bedroom and Living Room, with F1-scores of 81.5\% and 87.7\%, respectively. This suggests that the model effectively leverages long-range context to capture coherent layouts of major living spaces. In contrast, performance on smaller regions such as Kitchen and Bathroom is lower (around 65\%–70\%). This is partly due to the small spatial footprint of these classes. Even minor prediction errors can significantly affect pixel-based metrics, which is a common limitation in semantic segmentation evaluation. In addition, our inspection reveals occasional inconsistencies in the RPLAN annotations for small functional areas, which may further influence the quantitative scores. Nonetheless, the predictions often align better with plausible spatial priors than with raw labels. This highlights the model’s ability to infer reasonable structures even under imperfect supervision.

\begin{figure*}[ht]
  \centering
  \includegraphics[width=1\textwidth]{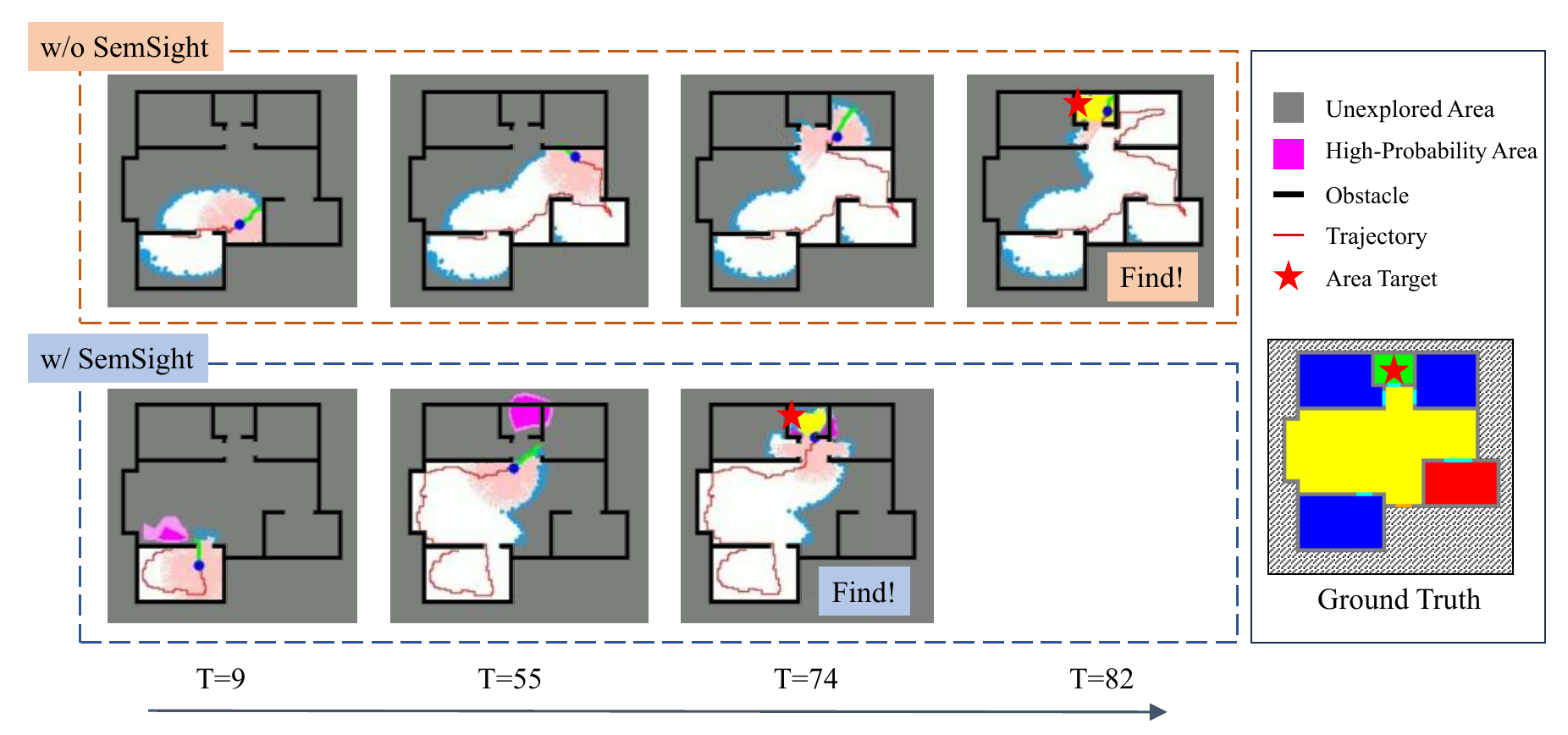} 
  \caption{\textbf{Qualitative comparison of navigation to target functional regions under two strategies.} The top row (w/o SemSight) shows trajectories of the exploration without semantic prediction, where the bathroom is found after 82 steps. The bottom row (w/ SemSight) shows trajectories guided by SemSight predictions, where the agent successfully navigates to the bathroom in 74 steps.}
  \label{fig:figure6}
\end{figure*}

\subsection{Qualitative Results}

As shown in Fig. \ref{fig:figure5}, we present prediction results for three representative scenes, using the bedroom as the structural target category. The model not only predicts the probability distribution of the target class but also infers functional areas such as kitchens, bathrooms, and balconies. The predicted spatial arrangements are consistent with common residential layouts. Even under highly limited observations, the predictions remain coherent, capturing inter-room relationships as well as structural boundaries (e.g., walls and doors). These results demonstrate that the model integrates contextual cues and spatial priors to generate reasonable global layouts under partial observability.

To further validate the model’s ability to reason about the spatial distribution of target categories during exploration, we select a representative scenario with the bedroom as the target. We visualize the evolution of probability heatmaps from step 10 to step 173 as shown in Fig. \ref{fig:figure4}. This sequential process highlights the model’s dynamic reasoning and its convergence ability over time.

\subsection{Closed-Loop Navigation Experiments} To further validate the practical value of the SemSight model in robotic navigation, we conduct closed-loop navigation experiments on our simulated exploration dataset. The predictions from SemSight are directly integrated into the navigation process to guide the agent’s decision-making and path planning. The task is defined as navigating to a specified functional target area from a random initial position under partial observability. We compare two strategies: a baseline exploration method without semantic prediction, and an enhanced closed-loop method that incorporates predictions from SemSight. A qualitative comparison of navigation to target functional regions under the two strategies is presented in Fig. \ref{fig:figure6}.

Navigation performance is evaluated using three metrics: Steps, Exploration Ratio, and Success weighted by Path Length (SPL). To ensure fairness, both strategies use the same step length, with the robot moving the same pixel distance at each action. As shown in Table \ref{tab:navi}, SemSight significantly reduces the number of steps required to discover the specified functional regions by 41.7\% and lowers the exploration ratio by 16.8\%, indicating more efficient search. Exploration Ratio is defined as the proportion of the environment explored when the target area is found. A lower value means the agent explores less of the environment while still achieving the goal, which reflects reduced exploration of irrelevant areas and better efficiency. SPL also improves by 16.2\%, reflecting higher path efficiency. The relatively high SPL compared with prior ObjectNav is expected, since our task evaluates reaching and exploring functional regions rather than locating objects. Overall, these results demonstrate that incorporating SemSight predictions provides valuable guidance for improving navigation performance.

\vspace{-3pt}
\begin{table}[htbp]
\centering
\caption{Navigation performance under two exploration strategies}
\label{tab:navi}
\begin{tabular}{lccc}
\toprule
\textbf{Method} & \textbf{Steps $\downarrow$} & \textbf{Exploration Ratio (\%) $\downarrow$} & \textbf{SPL (\%) $\uparrow$} \\
\midrule
w/o SemSight & 57.8 & 17.9 & 53.7 \\
w/ SemSight  & \textbf{33.7} & \textbf{14.9} & \textbf{62.4} \\
\bottomrule
\end{tabular}
\end{table}

\section{CONCLUSIONS}

In this work, we addressed the challenge of predicting semantic layouts in unexplored regions for target-driven navigation. Our SemSight model infers structural scene layouts and target area distributions under partial observations. This provides effective semantic priors for autonomous navigation. In future work, we plan to deploy the framework on real robotic platforms for performance evaluation. Moreover, we aim to extend our study to outdoor environments, exploring the capability of probabilistic semantic prediction in more complex and large-scale scenarios.

\addtolength{\textheight}{-2cm}   





 \bibliographystyle{unsrt}
 \bibliography{root1}

\end{document}